%% file: main.tex
\documentclass[letterpaper, 10 pt, conference]{ieeeconf}

\IEEEoverridecommandlockouts
\newif\ifanonymous
\anonymousfalse

\usepackage{graphicx}
\usepackage{amsmath}
\usepackage{amssymb}
\usepackage{booktabs}
\usepackage{multirow}
\usepackage{xcolor}
\usepackage{siunitx}
\usepackage[nohyperlinks]{acronym}
\usepackage[
  pdfencoding=auto,
  psdextra,
  unicode=true
]{hyperref}

\usepackage{caption}
\usepackage{comment}
\usepackage{amsfonts}       %
\usepackage{nicefrac}       %
\usepackage{microtype}      %
\usepackage{xcolor}         %
\usepackage{setspace} %

\usepackage{graphicx}
\usepackage{subcaption}
\usepackage{numprint} %
\usepackage{courier} %
\usepackage{xspace} %
\usepackage{balance} %

\usepackage{stfloats}

\usepackage{cleveref}
\usepackage[shortcuts]{extdash} %

\newacro{charuco}[ChArUco]{Chessboard + ArUco}
\newacro{dlt}[DLT]{Direct Linear Transform}
\newacro{mdm}[MDM]{Motion Diffusion Model}
\newacro{mocap}[MoCap]{motion capture}
\newacro{pnp}[PnP]{Perspective-n-Point}

\input{figure_commands}

\hypersetup{pdfinfo={
Title={Perceiving and Predicting Dynamic Sports: Spatiotemporal Intelligence in Vision Models for Egocentric Tennis Video},
Author={Tudor Braicu*, Joseph DelPreto*, Jared Boyer*, Mónika Farsang*, Maximilian Stölzle*, Tim Niklas Seyde, Daniela Rus},
Subject={Perceiving and Predicting Dynamic Sports: Spatiotemporal Intelligence in Vision Models for Egocentric Tennis Video},
Keywords={Deep learning for visual perception; 
Datasets for robotic vision; 
Perception-action coupling; 
Egocentric video; 
Exocentric video;
Allocentric video;
Vision models; 
Vision-language models; 
VLMs; 
Dynamic sports; 
Robotics; 
Datasets; 
Human subjects; 
Human motion activity; },
CreationDate={D:20261027050000},
ModDate={D:20261027050000}
}}

\title{\LARGE \bf
Perceiving and Predicting Dynamic Sports:\\ Spatiotemporal Intelligence in Vision Models for Egocentric Tennis Video
}

\ifanonymous
    \author{Anonymous Authors}
\else
    \input{authors}

\fi

\begin{document}
\bstctlcite{BSTcontrol}

\maketitle
\pagestyle{empty}
\thispagestyle{empty}

\begin{figure*}[!b]
    \centering
    \captionsetup{skip=6pt}
    \includegraphics[width=0.95\linewidth]{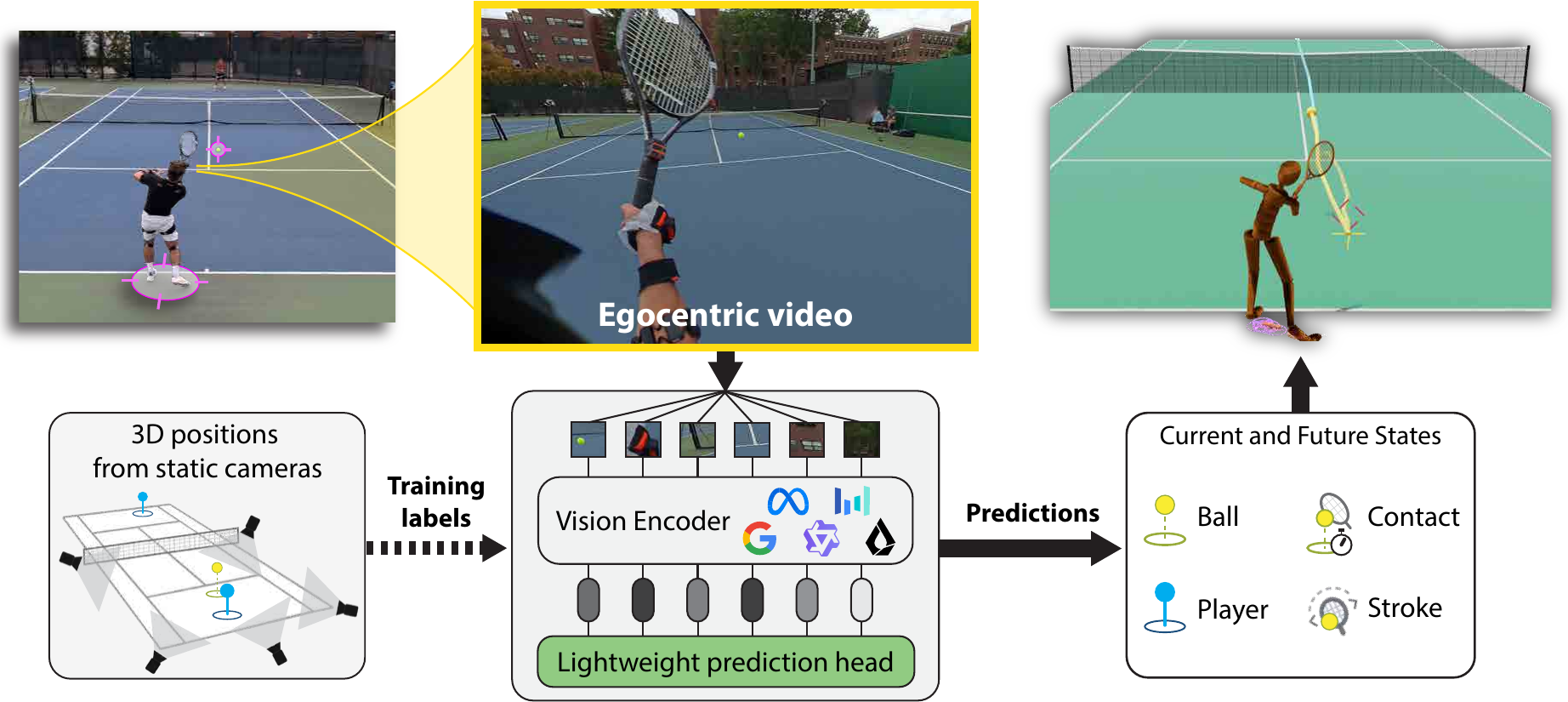}
    \caption{Visual models are trained on egocentric tennis videos to investigate spatiotemporal understanding, forecasting, and anticipation. The dataset will be publicly released, including synchronized allocentric videos and spatial and shot annotations.}
    \fglabel{overview}
\end{figure*}

\input{sections/00_abstract}
\input{sections/01_introduction}

\input{sections/04_dataset}

\input{sections/05_method}

\input{sections/06_experiments}
\input{sections/07_conclusion}

\ifanonymous\else
    \input{sections/08_acknowledgments}
\fi

\balance
\bibliographystyle{IEEEtran}
\bibliography{references_shorter}

\end{document}

%% file: figure_commands.tex
\newcommand{\fglabel}[1] {\label{fg:#1}}
\newcommand{\fgref}[1] {Figure~\ref{fg:#1}}

%% file: authors.tex
\author{%
    Tudor Braicu$^{*,1,2}$, Joseph DelPreto$^{*,1}$, Jared Boyer$^{*,1}$,
    M\'onika Farsang$^{*,1,3}$, Maximilian St\"olzle$^{*,1}$,\\
    Tim Niklas Seyde$^{1,4}$, and Daniela Rus$^{1}$%
    \thanks{\raggedright\hspace*{-1em}$^{*}$\,These authors contributed equally to this work.\protect\newline
        Corresponding author: Joseph DelPreto.}%
    \thanks{\raggedright\hspace*{-1em}%
        {\spaceskip=1.5pt $^{1}$\,Massachusetts Institute of Technology (MIT), Cambridge, MA 02139 USA.}\protect\newline
        $^{2}$\,Cornell University, Ithaca, NY 14853 USA.\protect\newline
        $^{3}$\,Technische Universit\"at Wien (TU Wien), 1040 Vienna, Austria.\protect\newline
        $^{4}$\,Liquid AI, Cambridge, MA 02142 USA.}%
    \thanks{\raggedright\hspace*{-1em}\protect\hypersetup{hidelinks}%
        \protect\settowidth{\hangindent}{E-mail:\ }%
        \advance\hangindent by\scriptspace\hangafter=1
        E-mail:
        \mbox{\{\href{mailto:delpreto@csail.mit.edu}{delpreto},\,%
          \href{mailto:rus@csail.mit.edu}{rus}\}@csail.mit.edu};\protect\newline
        \mbox{\{\href{mailto:jaredb@mit.edu}{jaredb},\,%
          \href{mailto:mstolzle@mit.edu}{mstolzle},\,%
          \href{mailto:tseyde@mit.edu}{tseyde}\}@mit.edu};
        \href{mailto:tb574@cornell.edu}{tb574@cornell.edu};
        \href{mailto:monika.farsang@tuwien.ac.at}{monika.farsang@tuwien.ac.at}.}%
    \thanks{\raggedright\hspace*{-1em}\textbf{Project website and video:} \protect\href{https://tennis-vision.csail.mit.edu}{https://tennis-vision.csail.mit.edu}}%
}

%% file: sections/00_abstract.tex
\begin{abstract}
Agile robots require spatiotemporal intelligence to interpret scenes, reason about interactions, and anticipate future events.  Yet it remains unclear how well foundation vision models support these capabilities, particularly from a moving egocentric camera, which is a natural view for robots.  To investigate this, we collect an annotated tennis dataset and train a suite of vision models on spatiotemporal tasks.  Our dataset contains egocentric video synchronized with five allocentric videos, estimated global positions, labeled contact times, and annotated stroke types.  We then evaluate seven prediction tasks from egocentric images: 2D ego player and 3D ball localization, future player and ball trajectories, forehand or backhand stroke anticipation, time to the next ego player ball contact, and 3D ball position at that contact.  We compare image encoders, video encoders, and vision-language models with jointly trained prediction heads.  The best performing models achieve mean current-frame localization errors of \qty{0.14}{m} for the ego player and \qty{0.54}{m} for the ball, future forecasting errors of \qty{0.34}{m} for the ego player and \qty{0.91}{m} for the ball over \qty{1.1}{s}, and a mean contact-time prediction error of \qty{0.14}{s}.  Auxiliary depth information substantially improves 3D ball localization, suggesting that inferring metric depth remains a challenge.  
Longer image histories do not improve ball forecasting, indicating that representations exploit scene context rather than rely solely on inter-frame motion.  
Finally, we propose a conditional diffusion model to generate whole-body kinematic tennis motion from vision encoder predictions on egocentric video at real-time rates, adapting movements as predictions are refined.  
These results move towards a deeper understanding of how foundation vision models can support real-time dynamic robots using egocentric video.
\end{abstract}

%% file: sections/01_introduction.tex
\begingroup
\setstretch{0.99}
\section{Introduction}

Advances in robot hardware and learned controllers are enabling increasingly agile whole-body movements~\cite{he2025asap} and athletic behaviors~\cite{haarnoja2024soccer,zhang2026latent,huang2026adapt}, but coordinating these movements with fast objects and other agents requires visual processing to be sufficiently accurate and timely to guide immediate action.  This motivates spatial intelligence that supports both estimating the current state of the environment and anticipating its evolution~\cite{zaidi2023esther,chen2024spatialvlm,zhou2025vlm4d}.

Spatial intelligence of vision-language models has been studied with question-answering techniques that probe their understanding of where objects and agents are located and how they are oriented~\cite{chen2024spatialvlm,cheng2024spatialrgpt}.
Accounting for how these relations evolve over time extends the scope to \emph{spatiotemporal intelligence}~\cite{zhou2025vlm4d}.
We further distinguish \emph{understanding} from \emph{forecasting}~\cite{grauman2022ego4d} -- understanding seeks to infer current spatial configurations, motions, and action cues from observed inputs, while forecasting aims to predict future states or actions using current and past observations.  Representations that expose this information can support downstream visuomotor control and embodied planning~\cite{nair2023r3m,driess2023palme}, while learned world models can predict future representations to guide action~\cite{zhou2025dinowm,assran2025vjepa2}. These applications motivate evaluating what information is accessible in large-scale pretrained visual representations and how adaptation affects their usefulness~\cite{majumdar2023vc1}.

We use tennis as a concrete setting to explore perceptual demands of highly dynamic sports: localizing players and fast-moving objects, anticipating trajectories, and coordinating body motion for imminent  contacts~\cite{zaidi2023esther,alberico2025egopingpong,ren2026smash}. Returning a tennis ball couples these capabilities -- spatial accuracy and timing are crucial since the players and ball continue moving during perception, planning, and execution~\cite{xiao2024ball,huang2026adapt}.  

There have been three main data modalities used by robots tackling these challenges. First, \emph{dedicated spatial sensing} uses \ac{mocap} sensors or calibrated camera systems to directly estimate player and ball positions for trajectory forecasting and control~\cite{zhang2026latent,huang2026adapt,zaidi2023esther,xiao2024ball}.  Second, \emph{allocentric videos}, recorded by external cameras at fixed court locations, support ball tracking and trajectory analysis~\cite{huang2019tracknet,dong2026racketvision}, player behavior modeling, simulated skill learning~\cite{zhang2021vid2player,zhang2023vid2player3d}, and spatial question-answering with foundation vision-language models~\cite{yang2026courtsi}. Third, \emph{egocentric perception}, recorded from a player-worn camera, has addressed image-space instance segmentation, forehand action phases, and pose estimation of other players~\cite{lee2024arsports,xue2023ae2,khirodkar2023egohumans}. Egocentric cues such as gaze \cite{shinkai2022gaze} and head orientation may also reveal aspects of attention and strategy that are less apparent from allocentric views.

For autonomous dynamic robots, egocentric video is especially relevant because it relies solely on a body-mounted camera rather than external sensors -- it thus removes workspace limitations and infrastructure demands.  However, this viewpoint presents a demanding test of perception: body and head motion mix camera and object motion, court landmarks and relevant objects leave the field of view, and dynamic objects such as a ball can appear small, blurred, or partially occluded.
The joint ability of foundation visual representations to support metric player and object state estimation and causal forecasting throughout such egocentric video of dynamic tasks remains insufficiently characterized.

We therefore ask: \emph{How well do foundation vision models support spatial understanding and forecasting from dynamic egocentric video?}

To support this evaluation, we collect a tennis dataset with synchronized video from a head-worn egocentric camera and five allocentric cameras. We annotate this data with segmentations and localization information via a custom pipeline that fuses automated inferences and manual labeling.

We then use the egocentric frames as model inputs and evaluate seven spatiotemporal understanding and forecasting tasks split into four groups. \emph{Spatial state estimation} predicts the current 2D ego player position and 3D ball position. \emph{Spatial state forecasting} predicts the future 2D ego motion and 3D ball motion. \emph{Stroke anticipation} predicts whether the ego player's next stroke will be forehand or backhand. \emph{Contact anticipation} predicts the remaining time until the ego player's racket-ball contact and the 3D ball position at contact. Together, these tasks probe the spatial and temporal understanding of pre-trained visual representations. 

We select 13 vision encoders spanning image encoders, video encoders, and vision-language models (VLMs).  We attach lightweight prediction heads to each encoder's output visual token stream, train the encoder and probes jointly on all tasks, and evaluate prediction accuracy on held-out rallies. We consistently observe 3D ball prediction errors and 2D ego player localization errors that are on par with expected ball motion across 1--2 frames and expected player motion across 3--7 frames, indicating promising dynamic performance. 

We perform ablations to investigate representations, dynamic information, and horizon lengths.  Ball prediction errors reduce significantly with supplementary depth information, indicating that metric depth estimation is weakly encoded in the visual representations. Longer conditioning histories do not improve 3D position inference; in fact, image encoders achieve identical ball forecasting performance with only one conditioning frame, relying instead on other spatial correlations to predict future motion. We study the impact of rally phase and horizon length, finding that higher temporal proximity to ego--ball contact time reduces ball localization error but increases ego player localization error, and that ball prediction error increases with the forecasting horizon.

Finally, we connect egocentric video understanding to whole-body tennis motion generation through a conditional motion diffusion model~\cite{tevet2023human}. Trained on our dataset, the diffusion model uses the above vision-based predictions of racket--ball contact position, time remaining until contact, and whether the ball is approaching or departing (derived from ball forecasting) as conditioning information. As new visual observations refine contact time and position estimates, the model adaptively re-plans its whole-body kinematic motion. The planning system is instantiated with a finetuned vision-only variant of the LFM2.5-VL-450M~\cite{liquidai2025lfm2}, which is the fastest of our evaluated VLM-derived models and capable of \qty{100}{Hz} inference on modern desktop GPUs. This pipeline demonstrates the utility of pre-trained spatiotemporal representations for downstream real-time robot applications.

Key contributions are as follows:
\begin{itemize}
    \item A tennis video dataset of \num{85} rallies with \num{53155} egocentric frames and five synchronized allocentric views, annotated with 3D ball trajectories, 2D court-frame player trajectories, stroke type labels, and contact times, to be released upon acceptance of the paper.
    \item A systematic evaluation of foundation vision model representations for seven tasks of spatiotemporal understanding and forecasting from egocentric tennis videos.
    \item Empirical findings that (i) spatiotemporal task performance is comparable across image encoders and VLMs, but not the video encoder; (ii) supplementary depth information reduces ball localization error by 50--60\%; (iii) longer conditioning histories yield little benefit.
    \item A conditional diffusion model that generates whole-body kinematic tennis motion conditioned on predictions from a vision encoder trained on egocentric video.
\end{itemize}
\endgroup

%% file: sections/04_dataset.tex
\section{Dataset Curation and Annotation Pipeline}

\subsection{Synchronized Egocentric and Allocentric Videos}

To collect a detailed video dataset of tennis play, we recorded an expert tennis player using a head-mounted egocentric camera and five static allocentric cameras in an outdoor setting.  Four static cameras were placed at eye level at each corner of the half-court on which the ego player was playing, and one static camera raised high above the ground was centered behind the court; the locations are illustrated in \fgref{data_court_distributions}.  The egocentric camera yielded video at \num{29.97} frames per second with resolution \num{3840}\,$\times$\,\num{2160} pixels.  The allocentric cameras yielded video at \num{29.97} or \num{59.94} frames per second with resolution at least \num{5120}\,$\times$\,\num{2880} pixels. 

All cameras recorded continuously for approximately \num{80}\,minutes. Manual annotations then marked the start and end times of each rally, and the contact times of each shot.  After extracting only the active times within rallies, the dataset spans approximately \num{29.6}\,minutes (\num{53155} egocentric frames) across \num{85} rallies.  Each rally is segmented from the opening shot to its annotated end: the final racket-ball contact, second bounce, or net contact. The rallies included \num{557} shots by the ego player and \num{574} shots by the opponent. The ego player's shots included \num{328} forehands and \num{229} backhands.

The rallies span both free play (\num{29} rallies) and structured practice (\num{56} rallies), with the latter including four different patterns designed to capture diverse ego player movement across the court. The resulting distributions of the ball and player locations within frames of the extracted rallies are illustrated in \fgref{data_court_distributions}. 

The data collection experiment was approved by the local branch of the Institutional Review Board (IRB).

\begin{figure}[tb]
    \centering
    \includegraphics[width=0.95\linewidth]{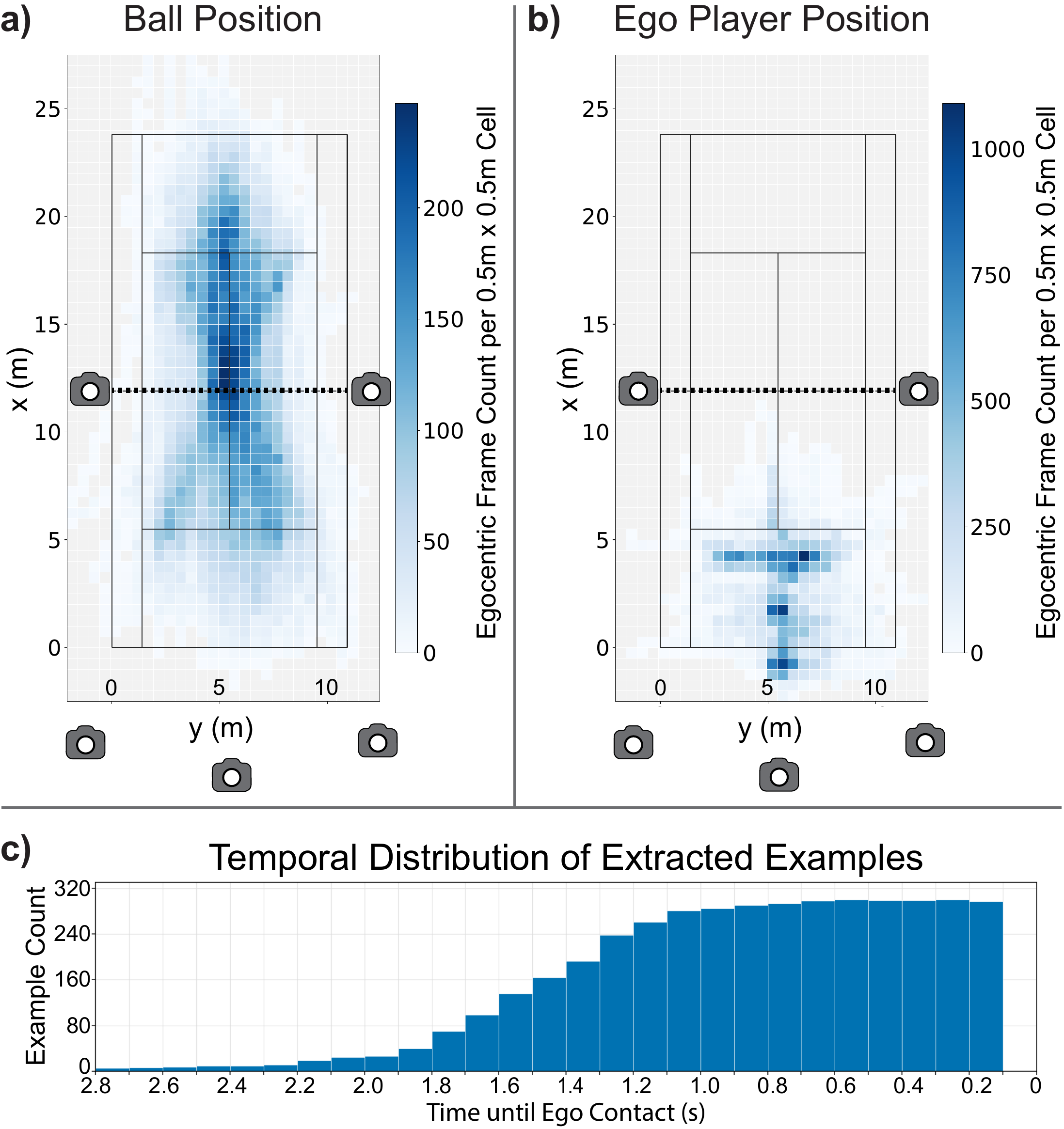}
    \caption{The dataset includes \num{53155} frames over \num{85} rallies.  Spatial distributions of samples are shown for (a) the ball and (b) the ego player.  The temporal distribution of samples relative to the ego player ball contact is shown in (c).  Locations of the five static cameras are depicted in (a) and (b).}
    \label{fg:data_court_distributions}
\end{figure}

\subsection{Camera Synchronization and Calibration}

At the beginning and end of the session, the ego player clapped three times. This synchronization signal, visible in all videos, was used in post-processing for manual alignment.

Calibration was performed in post-processing with a pinhole camera model, since all cameras use a linear field of view. Camera intrinsics are recovered from a \ac{charuco} calibration board that was held in front of allocentric cameras at the start of the session, and extrinsics are solved via \ac{pnp} with labeled court-corner correspondences, using OpenCV~\cite{bradski2000opencv}.

\subsection{State Estimation and Annotation}

We annotate recorded videos with information about the physical state and the upcoming play strategy.  This includes the ball position, the ego player position, and whether the ego player will next make a forehand or backhand stroke.

\subsubsection{Ball and Player Position Estimation}

We use the three allocentric cameras at the back of the court to estimate the ball and ego player positions. We first annotate the videos with ball and player pixel locations; after downscaling to $1280 \times 720$ resolution, we use an off-the-shelf tennis ball tracker~\cite{tarashima2023wasb} and a general-purpose object detector~\cite{yolo11} to obtain initial per-frame 2D anchors for the ball and ego player, which we then manually review and correct. 
Manual corrections were required for the ball and player annotations in approximately 30.5\% and 5.8\% of frames, respectively. 

We then fuse the pixel annotations from multiple cameras by triangulating the three calibrated and corrected 2D views using a \ac{dlt} solver to produce 3D ball and ego player position labels. The ball is triangulated freely in 3D, whereas the ego player localization uses a planar constraint to recover the foot-ground contact point on the 2D court plane, approximated as the bottom-center of the detected bounding box. At the downscaled $1280 \times 720$ resolution, median reprojection errors across the three views span 2.1--4.4$\,$px for the ball and  3.3--9.7$\,$px for the player. Ball and player positions are available for \qty{99.9}{\percent} and \qty{100}{\percent} of frames, respectively; triangulation requires valid 2D detections in at least two views, otherwise the frame is left unlabeled.

We also annotate the ball's 2D position in the egocentric view. The resulting egocentric ball position is available for 97.4\% of frames; the remainder are occlusions.

\subsubsection{Stroke Type Annotations}

To provide information about play strategy and motion prediction, we annotate the time of each ego player ball contact and whether it was a \emph{forehand} or \emph{backhand}. We also record the time of each opponent contact.%

%% file: sections/05_method.tex
\section{Vision Foundation Model Representations}
\label{lab:foundational_vision_rep} %

\subsection{Vision Models}~\label{lab:vision_models}
We evaluate a diverse set of vision models spanning image encoders, video encoders, and VLMs to assess how pretrained visual representations support egocentric tennis prediction.

\subsubsection{Image Encoders}
We consider four image encoders selected for their ubiquity and comparatively distinct training paradigms: DINOv3~\cite{simeoni2026dinov} is self-supervised, SigLIP~2~\cite{tschannen2025siglip} is language-supervised with contrastive and captioning objectives and is the vision tower of many VLMs, Depth Anything~3 (DA3)~\cite{lin2026depth} post-trains a DINOv2 backbone with dense depth supervision, and SAM~3~\cite{carion2026sam} post-trains a Perception Encoder~\cite{bolya2025PerceptionEncoder} backbone with dense segmentation supervision. We use checkpoints of comparable capacity: DINOv3 ViT-L/16 (300M), SigLIP~2 So400M/16 (400M), DA3-L (ViT-L/14, 300M), and SAM~3 (PE ViT-L+/14, 450M), i.e. ViT-L-class backbones~\cite{dosovitskiy2020vit} with 300–450M parameters.

\subsubsection{Video Encoders}
V-JEPA 2~\cite{assran2025vjepa2} is a video encoder self-supervised to predict masked latent representations. We use the ViT-L/16 checkpoint (300M) with $2 \times 16 \times 16$ tubelet patches, matching the capacity of the image encoders for multi-frame processing.

\subsubsection{Vision-Language Models}
We consider three VLMs at the 3--4B scale: Qwen3-VL-4B~\cite{qwen3vl2025}, LFM2.5-VL-3B~\cite{liquidai2025lfm2}, and PLM-3B~\cite{cho2025PerceptionLM}, whose vision towers derive from SigLIP~2-Large (300M), SigLIP~2 So400m NaFlex (400M), and the Perception Encoder~\cite{bolya2025PerceptionEncoder}, respectively. Because the encoders differ, we isolate the contribution of the language backbone and multimodal training by evaluating each VLM both as a vision tower alone and with its full language stack. We additionally include LFM2.5-VL-450M, whose 86M-parameter vision tower is several times smaller than those of the larger VLMs, as a compute-constrained variant to enable downstream real-time planning experiments  (Section~\ref{sec:mdm}).

\subsection{Backbone Adaptation}
\label{lab:backbone_adaptation}

\subsubsection{Prediction Head}
To adapt visual representations with minimal computational overhead, we attach task-specific prediction heads to the encoder's output token stream (patch embeddings and, for VLMs, language tokens). Each head applies cross-attention from learnable queries (one per output, $N$ for $N$-step forecasts) to the token stream, followed by a 2-layer MLP. All heads are trained jointly on a shared encoder and share an additive positional embedding over the tapped tokens. We use an L2 loss (weight 1.0) for continuous targets and a cross-entropy loss (weight 0.1) for stroke type.

\subsubsection{Fine-tuning and Training Details}
All experiments use a batch size of 32 and 30 training epochs. We use a learning rate of $10^{-4}$ for the newly initialized prediction heads and $10^{-5}$ for the pretrained vision backbones. For vision-language models, the language component is finetuned using Low-Rank Adaptation (LoRA)~\cite{hu2021lora} rather than full-parameter fine-tuning. Thus, our fine-tuning protocol involves up to three components: a fully trained prediction head, a vision backbone finetuned at a reduced learning rate, and a parameter-efficient LoRA adaptation of the language model.

\subsubsection{Vision Tower and Vision-Language Settings}
\label{sec:vlm_settings}
We evaluate VLMs in two settings. In the \emph{vision-tower} setting, we retain only the vision tower, using its post-projector tokens as the representation stream. In the \emph{vision-language} setting, we include both the vision and language components. In line with practices in VLM fine-tuning for auxiliary tasks~\cite{reuss2025flower}, we tap representations located 50\% through language tower depth and discard the second half. For all VLMs in the \emph{vision-language} setting, we inject a standardized text prompt after the encoded image tokens to describe the overall task goal.

%% file: sections/06_experiments.tex
\section{Experimental Results and Discussion}

\subsection{Experimental Setup}

We evaluate all models on the same fixed data splits. Rallies are assigned wholly to one split, so that no temporal window crosses a rally boundary. Splits are stratified across each of the drill sub-types and free-play, to avoid rally type imbalances. The realized split ratios are \qty{63.1}{\percent} train, \qty{17.7}{\percent} validation, \qty{19.2}{\percent} test, corresponding to \num{53} / \num{16} / \num{16} rallies. Training occurs on NVIDIA B200 and H200 GPUs; individual run durations vary between 4--24 hours depending on the model.

\subsection{Multi-Task Prediction}

Each model receives two images as input: the current ``anchor'' frame and one preceding frame 0.1\,s (3 frames) earlier. We jointly optimize seven prediction heads covering five targets. \emph{3D ball position} and \emph{2D player position} are each predicted at the anchor frame and over a horizon of 12 future frames spanning 33 time steps (1.10\,s). \emph{Shot type} classifies the upcoming stroke as forehand or backhand. \emph{Time to contact} predicts the time until the ego player's next ball contact. \emph{Ball position at contact} predicts 3D ball position at that time.

\begin{figure}[t]
  \centering
  \includegraphics[width=0.99\columnwidth]{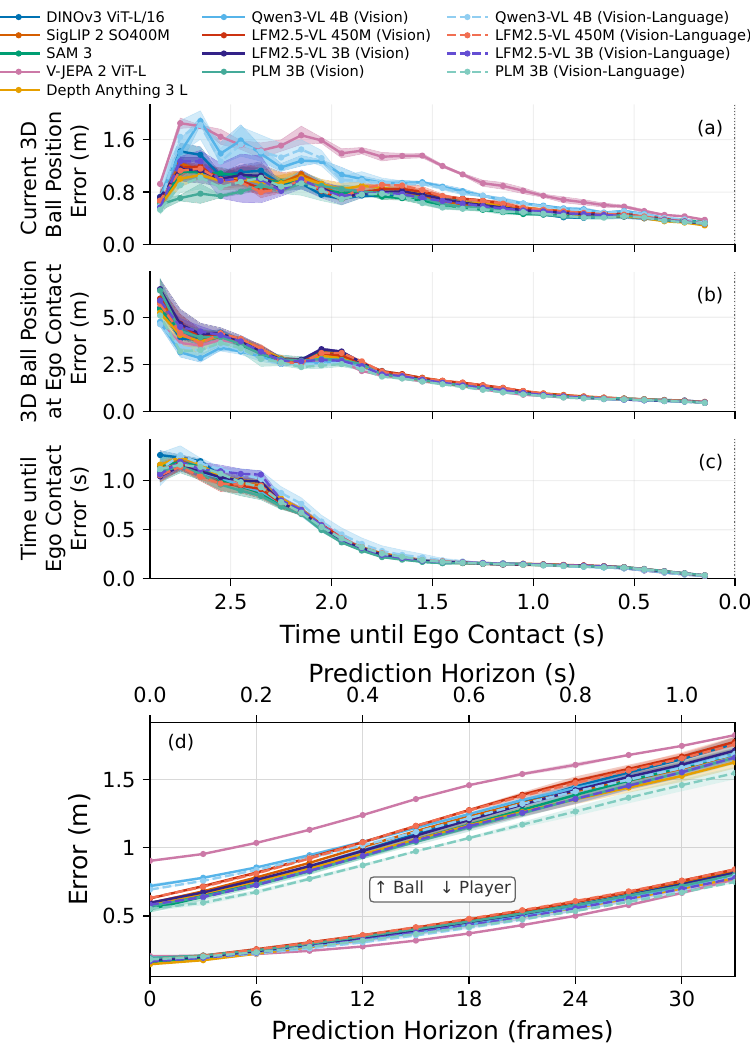}
  \caption{
  Prediction errors on approaching test-split shots versus time to contact: (a)~current 3D ball position, (b)~ball position at contact, and (c)~time to contact. Lines show means over \qty{0.1}{\second} intervals; shading shows $\pm1$ SD over three seeds. (d) Ball 3D and player 2D future predictions over a time horizon of $\sim$1.10\,s, across all models, averaged over three seeds.
  }
  \label{fig:vision-performance-time-to-contact}
\end{figure}

Table~\ref{tab:results} reports test-split metrics for all 13 configurations. Current-frame 3D ball errors range from $0.54$\,m (PLM~3B) to $0.90$\,m (V-JEPA~2). VLMs and vision towers fall within a narrower band ($0.54$ to $0.71$\,m, with Qwen3-VL highest), and image encoders are the most tightly clustered ($0.55$ to $0.60$\,m). Current-frame 2D player errors vary less across models ($0.14$ to $0.18$\,m), with DA3 lowest. Forecasting separates the models more clearly: PLM~3B attains the lowest future ball error ($0.91$\,m) and V-JEPA~2 by far the highest ($1.27$\,m), yet V-JEPA~2 attains the lowest future player error ($0.34$\,m) and retains this advantage across the full horizon, while its ball error remains the highest throughout (Fig.~\ref{fig:vision-performance-time-to-contact} (d)). Shot-type accuracy is near-saturated for all models ($98.2$ to $99.9\%$), and time-to-contact errors fall within a narrow range of $141$ to $149$\,ms; Fig.~\ref{fig:vision-performance-time-to-contact} shows how ball and contact errors evolve as contact approaches. Adding the language stack yields small gains for the three larger VLMs, most notably for PLM~3B on ball-related tasks. Overall, image encoders and VLMs perform comparably; PLM~3B leads on ball tasks and, together with LFM2.5-VL~3B, SAM 3, and DA3, is the most consistently performant across spatial tasks, while Qwen3-VL trails slightly on ball localization and V-JEPA~2 is weakest on ball localization but strongest on player forecasting.

\begin{table*}[t]
  \centering
  \small
  \setlength{\tabcolsep}{5pt}
  \renewcommand{\arraystretch}{1.15}
    \caption{Prediction accuracy across probed visual representations. Task prediction heads are trained jointly on a shared encoder. ``Current'' predictions are at the final conditioning frame; ``future'' predictions are the mean over 12 subsequent frames with stride 0.1\,s. Results reflect performance on the test split, using model checkpoints selected based on the validation split, across three seeds.}
  \begin{tabular*}{\textwidth}{@{\extracolsep{\fill}}l cc cc ccc@{}}
    \toprule
    \textbf{Model} & \multicolumn{2}{c}{\textbf{3D Ball pos.}} & \multicolumn{2}{c}{\textbf{2D Player pos.}}
                   & \textbf{Shot type} & \textbf{Time to} & \textbf{Ball pos. at} \\
    \cmidrule(lr){2-3} \cmidrule(lr){4-5}
                  & Current (m) $\downarrow$ & Future (m) $\downarrow$ & Current (m) $\downarrow$ & Future (m) $\downarrow$ & (acc.\ \%) $\uparrow$ & \textbf{contact} (ms) $\downarrow$& \textbf{contact} (m) $\downarrow$ \\
    
        \midrule
    \multicolumn{8}{@{}l}{\textit{Image encoders}} \\
    \texttt{DINOv3}    & $0.58 \pm 0.01$ & $1.02 \pm 0.01$ & $0.16 \pm 0.01$ & $0.38 \pm 0.01$ & $98.7 \pm 0.0$ & $\mathbf{141.3 \pm 1.6}$ & $1.03 \pm 0.01$ \\
    \texttt{SigLIP 2}    & $0.60 \pm 0.01$ & $1.03 \pm 0.00$ & $0.16 \pm 0.00$ & $0.39 \pm 0.00$ & $98.4 \pm 0.2$ & $148.8 \pm 1.6$ & $1.08 \pm 0.02$ \\
    \texttt{SAM 3}              & $0.55 \pm 0.03$ & $0.99 \pm 0.03$ & $0.15 \pm 0.00$ & $0.38 \pm 0.00$ & $98.4 \pm 0.3$ & $142.7 \pm 3.1$ & $1.01 \pm 0.01$ \\
    \texttt{DA3} & $0.58 \pm 0.01$ & $0.99 \pm 0.02$ & $\mathbf{0.14 \pm 0.01}$ & $0.36 \pm 0.01$ & $98.4 \pm 0.0$ & $142.8 \pm 1.6$ & $1.00 \pm 0.02$ \\
    \cmidrule[0.2pt](lr){1-8}
    \multicolumn{8}{@{}l}{\textit{Video encoders}} \\
    \texttt{V-JEPA 2}     & $0.90 \pm 0.00$ & $1.27 \pm 0.00$ & $0.18 \pm 0.00$ & $\mathbf{0.34 \pm 0.01}$ & $\mathbf{99.9 \pm 0.1}$ & $144.0 \pm 0.1$ & $0.97 \pm 0.01$ \\
    \cmidrule[0.2pt](lr){1-8}
    \multicolumn{8}{@{}l}{\textit{Vision towers}} \\
    \texttt{Qwen3-VL 4B}        & $0.71 \pm 0.01$ & $1.08 \pm 0.01$ & $0.18 \pm 0.01$ & $0.37 \pm 0.00$ & $\mathbf{99.9 \pm 0.1}$ & $142.5 \pm 3.1$ & $1.00 \pm 0.01$ \\
    \texttt{LFM2.5-VL 450M}     & $0.63 \pm 0.00$ & $1.08 \pm 0.01$ & $0.17 \pm 0.00$ & $0.40 \pm 0.01$ & $98.3 \pm 0.2$ & $146.3 \pm 2.1$ & $1.07 \pm 0.01$ \\
    \texttt{LFM2.5-VL 3B}       & $0.59 \pm 0.01$ & $1.02 \pm 0.01$ & $0.16 \pm 0.00$ & $0.38 \pm 0.00$ & $98.2 \pm 0.1$ & $145.2 \pm 0.7$ & $1.03 \pm 0.00$ \\
    \texttt{PLM 3B}             & $0.56 \pm 0.01$ & $0.98 \pm 0.02$ & $0.18 \pm 0.01$ & $0.39 \pm 0.01$ & $98.8 \pm 0.4$ & $143.7 \pm 0.6$ & $1.02 \pm 0.02$ \\
    \cmidrule[0.2pt](lr){1-8}
    \multicolumn{8}{@{}l}{\textit{Vision--language models}} \\
    \texttt{Qwen3-VL 4B}        & $0.69 \pm 0.01$ & $1.06 \pm 0.01$ & $0.17 \pm 0.00$ & $0.36 \pm 0.01$ & $99.8 \pm 0.1$ & $147.3 \pm 5.8$ & $0.98 \pm 0.02$ \\
    \texttt{LFM2.5-VL 450M}     & $0.63 \pm 0.01$ & $1.07 \pm 0.01$ & $0.17 \pm 0.01$ & $0.40 \pm 0.01$ & $98.2 \pm 0.3$ & $146.9 \pm 1.5$ & $1.08 \pm 0.01$ \\
    \texttt{LFM2.5-VL 3B}       & $0.57 \pm 0.01$ & $0.98 \pm 0.01$ & $0.16 \pm 0.00$ & $0.36 \pm 0.01$ & $99.4 \pm 0.2$ & $146.7 \pm 1.1$ & $1.00 \pm 0.02$ \\
    \texttt{PLM 3B}             & $\mathbf{0.54 \pm 0.02}$ & $\mathbf{0.91 \pm 0.00}$ & $0.17 \pm 0.01$ & $0.36 \pm 0.00$ & $99.7 \pm 0.1$ & $144.4 \pm 1.4$ & $\mathbf{0.95 \pm 0.02}$ \\
    \bottomrule
  \end{tabular*}
  \label{tab:results}
\end{table*}

\subsection{Injecting Information to Probe Spatial Representations}
\label{sec:aux_info}

We study how providing auxiliary inputs from the annotations to the prediction head affects 3D ball localization, using SigLIP~2 finetuned with a single head for the anchor-frame ball position. We consider three inputs: the ball's 2D image position in the ego frame (\textit{ball}), the ego player's 2D court position (\textit{player}), and the planar ego-ball distance (\textit{depth}), each evaluated individually and in all pairwise and full combinations. Inputs are sinusoidally embedded and injected into the head's cross-attention block via adaptive layer normalization (AdaLN)~\cite{perez2018film}. As a control, we train an image-free MLP on the auxiliary inputs alone.

As shown in Table~\ref{tab:sideinfo}, \textit{depth} reduces error from \qty{0.51}{m} (image only) to \qty{0.21}{m}, and combining it with \textit{player} yields the lowest error (\qty{0.19}{m}). Yet \textit{ball} and \textit{player} alone provide no benefit, and adding \textit{ball} slightly degrades the combinations. This suggests that depth supplies geometric information hard to infer from images, but ball pixel position is redundant with visual representation. The image-free MLP also benefits most from \textit{depth}, but unlike SigLIP~2, it gains more from \textit{ball}, performing best with all three inputs (\qty{1.07}{m}). The substantially higher errors of image-free MLP variants show that the auxiliary inputs alone cannot replace visual data.

\begin{table}[t]
  \centering
  \small
  \setlength{\tabcolsep}{10pt}
  \renewcommand{\arraystretch}{1.15}
    \caption{
    \underline{Current} 3D ball position error with auxiliary inputs. Results reflect test-split errors over three seeds, using checkpoints selected based on the validation split. $^\dagger$Image-free baseline using only auxiliary inputs.
    }
  \begin{tabular*}{\columnwidth}{@{\extracolsep{\fill}}l cc@{}}
    \toprule
    \textbf{Auxiliary Information Added} & MLP$^\dagger$  $\downarrow$& \texttt{SigLIP 2} $\downarrow$\\
    \midrule
    None (image only)     & --                       & $0.51 \pm 0.01$                    \\
    Ball                  & $2.83 \pm 0.01$          & $0.54 \pm 0.00$                    \\
    Player                & $4.28 \pm 0.03$          & $0.51 \pm 0.01$                    \\
    Depth                 & $2.39 \pm 0.01$          & $0.21 \pm 0.01$                    \\
    Ball + player         & $2.66 \pm 0.03$          & $0.53 \pm 0.01$                    \\
    Ball + depth          & $2.12 \pm 0.01$          & $0.25 \pm 0.02$                    \\
    Player + depth        & $1.24 \pm 0.01$          & $\mathbf{0.19 \pm 0.01}$           \\
    Ball + player + depth & $\mathbf{1.07 \pm 0.02}$ & $0.23 \pm 0.03$                    \\
    \bottomrule
  \end{tabular*}

  \label{tab:sideinfo}
\end{table}

\begin{table}[t]
  \centering
  \footnotesize
  \setlength{\tabcolsep}{3pt}
  \renewcommand{\arraystretch}{1.15}
    \caption{
    \underline{Future} 3D ball position error versus input history (\emph{Frames}: number of input frames; \emph{Span}: history duration). Results reflect test-split errors over three seeds, using checkpoints selected based on the validation split. $^\dagger$Image-free baseline using only ground-truth 3D ball history.
    }
    \begin{tabular*}{\columnwidth}{@{\extracolsep{\fill}}cc ccc@{}}
    \toprule
    \textbf{Frames} & \textbf{Span} & GT MLP$^\dagger$ $\downarrow$ & \texttt{SigLIP 2} $\downarrow$ & \texttt{V-JEPA 2} $\downarrow$ \\
    \midrule
    1 & --     & $2.924 \pm 0.003$          & $\mathbf{1.02 \pm 0.02}$ & --                        \\
    2 & 100 ms & $0.297 \pm 0.001$          & $\mathbf{1.02 \pm 0.05}$ & $1.22 \pm 0.01$           \\
    2 & 300 ms & $0.257 \pm 0.004$          & $1.08 \pm 0.05$          & $1.25 \pm 0.01$           \\
    4 & 300 ms & $\mathbf{0.238 \pm 0.003}$ & $1.08 \pm 0.02$          & $\mathbf{1.18 \pm 0.02}$  \\
    \bottomrule
  \end{tabular*}
  \label{tab:history}
\end{table}

\subsection{Varying History to Probe Temporal Representations}

We study how probed visual representations encode temporal context through history conditioning by varying the number and spacing of input frames while predicting future 3D ball position. In Table~\ref{tab:history}, we finetune SigLIP 2 on four settings and V-JEPA 2 on three, omitting single-frame history since tubelets must be at least length 2. 

An image-free MLP trained on ground truth 3D ball positions shows the necessity of temporal information: from a single position it cannot forecast (\qty{2.92}{m} error), but a second position \qty{100}{ms} earlier suffices to estimate velocity and reduces error to \qty{0.30}{m}, with longer or denser history providing only modest further gains.

Image-based models show a different pattern. With only a single frame, SigLIP 2 already achieves \qty{1.02}{m}, far below the position-only baseline, suggesting that temporal information beyond the ball's instantaneous position is being extracted. However, additional frames provide no benefit (\qtyrange{1.02}{1.08}{m}), so the representation does not recover inter-frame ball displacement. This is consistent with the auxiliary information ablation of Section~\ref{sec:aux_info}: without accurate depth inference, finite differences yield noisy velocity estimates, and models may instead rely on spatial cues found within a single frame. V-JEPA 2 improves slightly with more frames (\qty{1.22}{m} to \qty{1.18}{m}), but remains worse than SigLIP~2 on a single frame.

\begin{figure}[tb]
    \centering
    \includegraphics[width=0.95\linewidth]{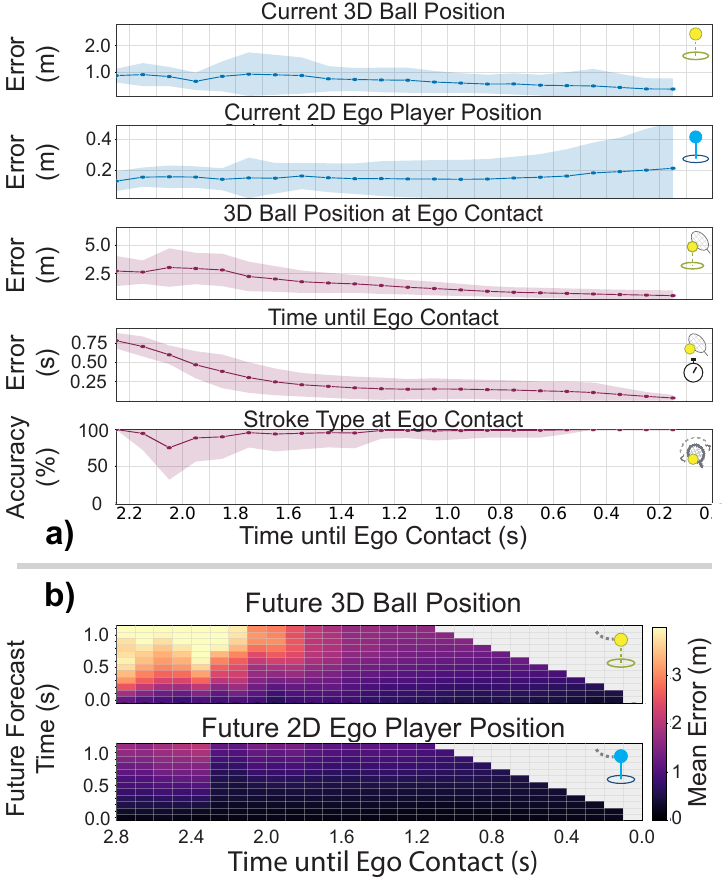}
    \caption{LFM2.5-VL 450M results on approaching test-split shots until contact. (a) Predictions for a single frame horizon tend to improve as the ball nears the ego player. (b) Forecasted predictions tend to improve as the ball nears the ego player and degrade as the future horizon increases.}
    \label{fg:lfm450_evaluation_static_and_future}
\end{figure}

\subsection{Effect of Backbone Adaptation}
We compare our training configuration on the LFM2.5-VL 3B vision-language model (finetuned vision encoder with LoRA-adapted language stack) against a fully frozen variant where the vision and the language towers are fixed and only the prediction heads are trained. We find that freezing degrades every task: the error of current ball position rises from $0.57$ to $0.96$\,m, and future ball position from $0.98$ to $1.47$\,m. The 2D player tasks are more robust, with current-frame error rising from $0.16$ to $0.18$\,m and future-horizon error from $0.36$ to $0.45$\,m. Shot-type classification falls from $99.4\%$ to $98.3\%$, time to contact error rises from $146.7$ to $162.0$\,ms, and ball position at contact error rises from $1.003$ to $1.225$\,m. These results suggest that the frozen features can already largely support current 2D player and shot type predictions, but do not encode precise visual cues needed to localize the ball in 3D from a moving egocentric view.

\subsection{Conditional Motion Diffusion for Kinematic Tennis Play}
\label{sec:mdm}

\begin{figure*}[tb]
    \centering
    \includegraphics[width=1\linewidth]{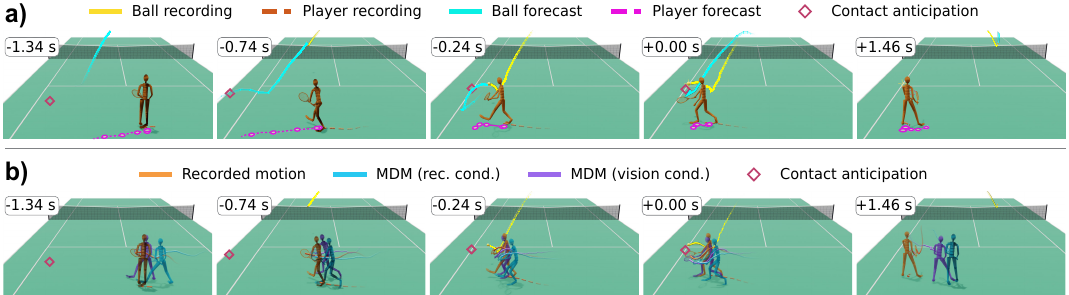}
    \caption{(a)~LFM2.5-VL-450M forecasts ball and player positions and anticipates contact; crosshairs locate the recorded contact. (b)~Recorded  ego motion versus MDM rollouts conditioned on annotations (``rec.\ cond.'') or vision predictions (``vision cond.''). Trails fade toward the past; ball forecasts fade toward the future. $t=0$ denotes the recorded racket--ball contact.}
    \label{fg:diffusion_shot_09_combined}
\end{figure*}

We demonstrate how foundation vision models' spatiotemporal understanding of egocentric video supports whole-body tennis motion generation. We use \acp{mdm}, a leading approach to conditional kinematic motion generation~\cite{tevet2023human,shafir2024prior,zhao2025dartcontrol}.

Alongside the videos, we recorded ego player motion using the wearable Xsens MTw Awinda system. We compute joint rotations from the relative orientations of adjacent body segments, using the calibrated T-pose as a reference. Motion frames contain \num{147} features: pelvis translation and 6D rotations for the pelvis and \num{23} joints, including the racket attachment. Each window's coordinate frame is centered on the ground below the pelvis and aligned with its heading at the last history frame. The conditioning vector is
\begin{equation}\small
    \begin{aligned}
        \mathbf c_t = [\tilde{\mathbf b}^{\,*}_t,\tau_t^{\mathrm{next}},m_t^{\mathrm{next}},
        \tau_t^{\mathrm{prev}},m_t^{\mathrm{prev}},\mathbf p^{xy}_t,\sin\psi_t,\cos\psi_t]\in\mathbb R^{11},
    \end{aligned}
\end{equation}
where $\tilde{\mathbf b}^{\,*}_t$ is the anticipated ball-contact position in this local frame, $\tau_t^{\mathrm{next}}$ is the time remaining until contact, and $\tau_t^{\mathrm{prev}}$ is the time elapsed since the previous player contact. Binary masks $m_t^{\mathrm{next}}$ and $m_t^{\mathrm{prev}}$ indicate whether the upcoming-contact inputs and previous-contact timing, respectively, are supplied or masked out. The global pelvis position $\mathbf p^{xy}_t$ and heading $\psi_t$ help learn to return to a neutral court pose after a shot. Training reconstructs recorded motion from noisy samples, conditioned on annotated tennis states, using the vision dataset's rally-based train / validation / test splits.

To obtain tennis conditions from video, we use the LFM2.5-VL-450M vision encoder and prediction heads with two recorded images approximately \qty{100}{ms} apart. Its compact 86M-parameter encoder enables real-time visual conditioning at \qty{100}{Hz} on an RTX~5090, while the model localizes the ball more accurately than the larger Qwen3-VL~4B vision tower and stays within \qty{5}{ms} of the best contact-time error (Table~\ref{tab:results}). Its predictive capabilities are displayed in~\fgref{lfm450_evaluation_static_and_future}. The predicted racket--ball contact position, transformed into the local pelvis frame, supplies $\tilde{\mathbf b}^{\,*}_t$, while the predicted time remaining to contact supplies $\tau_t^{\mathrm{next}}$. From current and forecast ball positions, we estimate velocity along the court to classify approaching shots (toward the ego player) and departing shots (away). We set $m_t^{\mathrm{next}}=1$ for approaching shots with a valid contact estimate; otherwise, $m_t^{\mathrm{next}}=0$ masks both target inputs. We recognize player contact when successive vision estimates show an approaching-to-departing reversal near the predicted player position. The time elapsed since the latest recognized contact supplies $\tau_t^{\mathrm{prev}}$, with $m_t^{\mathrm{prev}}=1$ after the first recognized contact and zero beforehand. %

We initialize each rollout with two recorded motion frames. The model predicts \num{24} frames at \qty{60}{Hz} (\qty{400}{ms}); we execute the first six (\qty{100}{ms}) and use the last two executed frames as history for the next prediction. Accumulated generated pelvis position and heading supply the global $\mathrm{SE}(2)$ condition and the local frame for contact targets. At each replanning step, we update the predicted contact position and time remaining, allowing the motion model to revise its motion as the ball approaches. This is motivated by the increasing accuracy of these vision estimates closer to contact (\Cref{fig:vision-performance-time-to-contact}). Throughout the rollout, the vision model receives successive frames from the original recorded video. Its observations therefore follow the recorded player's camera trajectory and do not change in response to the generated character's movement.

\fgref{diffusion_shot_09_combined} shows synchronized vision predictions and motion comparisons for one held-out test shot. \Cref{tab:rollout-evaluation} compares recorded motion with rollouts conditioned on annotations or vision predictions, using matching initial histories and random seeds on held-out sequences. We measure ball--racket distance at annotated contact and its minimum within $\pm\qty{100}{ms}$, together with errors in racket-center position, racket-plane orientation, and velocity relative to recorded motion.

\providecommand{\rolloutmetric}[3]{\shortstack[r]{\num{#1}\\[-1pt]{\scriptsize [\num{#2},\,\num{#3}]}}}
\begin{table}[t]
\centering\footnotesize
\setlength{\tabcolsep}{0.7pt}
\renewcommand{\arraystretch}{1.2}
\caption{
MDM rollouts on 16 test sequences: means over contacts with bootstrapped 95\% confidence intervals in brackets. Racket errors are relative to recorded ego player motion; minimum distance uses a $\pm$\qty{100}{ms} window.
}
\label{tab:rollout-evaluation}
\begin{tabular*}{\columnwidth}{@{\extracolsep{\fill}}lrrrrr@{}}
\toprule
& \multicolumn{2}{c}{\shortstack{Ball--racket\\proximity}} & \multicolumn{3}{c}{\shortstack{Error relative to\\recorded motion}} \\
\cmidrule(lr){2-3}\cmidrule(l){4-6}
Approach & \shortstack{At hit\\(m) $\downarrow$} & \shortstack{Min.\\(m) $\downarrow$} & \shortstack{Center\\(m) $\downarrow$} & \shortstack{Plane\\($^\circ$) $\downarrow$} & \shortstack{Velocity\\(m/s) $\downarrow$} \\
\midrule
\shortstack[l]{Recorded\\motion} & \rolloutmetric{0.89}{0.76}{1.02} & \rolloutmetric{0.81}{0.69}{0.92} & \num{0.00} & \num{0.00} & \num{0.00} \\
\midrule
\shortstack[l]{MDM\\(rec. cond.)} & \rolloutmetric{1.11}{1.04}{1.19} & \rolloutmetric{0.98}{0.93}{1.04} & \rolloutmetric{0.76}{0.68}{0.84} & \rolloutmetric{35.97}{33.10}{38.90} & \rolloutmetric{4.46}{3.91}{5.07} \\
\addlinespace[3pt]
\shortstack[l]{MDM\\(vision cond.)} & \rolloutmetric{1.28}{1.12}{1.42} & \rolloutmetric{1.17}{1.04}{1.33} & \rolloutmetric{0.95}{0.78}{1.10} & \rolloutmetric{41.31}{36.03}{46.43} & \rolloutmetric{5.38}{4.80}{5.92} \\
\bottomrule
\end{tabular*}
\end{table}

Results demonstrate that a VLM vision encoder finetuned for spatiotemporal prediction on egocentric tennis video can provide real-time conditioning for whole-body ego player motion generation. In the qualitative example (\fgref{diffusion_shot_09_combined}), the generated player approaches the predicted contact point with well-timed motion that anticipates racket--ball contact, while the contact estimate becomes more accurate as the ball nears.

A limitation is the quality of the reconstructed recordings: even recorded motion has a mean ball--racket distance of \qty{0.89}{m} at annotated contact (\Cref{tab:rollout-evaluation}). This mismatch limits both motion learning and evaluation. Precise racket--ball alignment remains unresolved and requires improvements in both recording quality and vision prediction accuracy. The demonstration also remains kinematic, without a policy for physical execution in simulation or reality.

%% file: sections/07_conclusion.tex
\section{Conclusion}
We present a systematic study of spatiotemporal understanding and forecasting from dynamic egocentric video, using tennis as a challenging setting for spatial perception and anticipation. Across seven tasks, we find that image encoders and VLMs provide broadly competitive performance and we analyze how their strengths differ by task.
Our ablations further show that providing additional depth information substantially improves ball localization, but that longer visual histories do not necessarily improve performance.

Building on these results, we present whole-body motion generation via conditional motion diffusion.  This demonstrates that the egocentric estimates could support future pipelines for real-time robot control. 

This paper takes a step towards a deeper understanding of how foundation vision models can support highly dynamic robotic tasks.  Tennis is one instance of tasks where a robot must maintain situational awareness and act under tight time constraints from a moving viewpoint; future work will test whether the identified representations and adaptation strategies transfer to other highly dynamic settings such as team sports, catching and throwing, and navigation among fast agents.

%% file: sections/08_acknowledgments.tex
\section*{Acknowledgments}
This work was supported in part by the GIST-MIT Research Collaboration grant funded by the Gwangju Institute of Science and Technology (GIST), Schmidt Sciences through the AI2050 program, the Boeing company, and the Department of the Air Force Artificial Intelligence Accelerator for research and computing resources under Cooperative Agreement No.~FA8750-19-2-1000. This work used computing resources provided by Advanced Research Computing at Hopkins (ARCH).

The views and conclusions expressed in this paper are those of the authors and do not necessarily reflect the official policies, whether express or implied, of the Department of the Air Force or the U.S. Government. The U.S. Government may reproduce and distribute reprints for government purposes regardless of any copyright notices.

We used AI agents to generate parts of our codebase, including Claude Sonnet 5, Claude Opus 5, Claude Fable 5.1, GPT-5.6 Sol, and GPT-6 Astra.

%% file: main.bbl
\begin{thebibliography}{10}
\providecommand{\url}[1]{#1}
\csname url@rmstyle\endcsname
\providecommand{\newblock}{\relax}
\providecommand{\bibinfo}[2]{#2}
\providecommand\BIBentrySTDinterwordspacing{\spaceskip=0pt\relax}
\providecommand\BIBentryALTinterwordstretchfactor{4}
\providecommand\BIBentryALTinterwordspacing{\spaceskip=\fontdimen2\font plus
\BIBentryALTinterwordstretchfactor\fontdimen3\font minus
  \fontdimen4\font\relax}
\providecommand\BIBforeignlanguage[2]{{%
\expandafter\ifx\csname l@#1\endcsname\relax
\typeout{** WARNING: IEEEtran.bst: No hyphenation pattern has been}%
\typeout{** loaded for the language `#1'. Using the pattern for}%
\typeout{** the default language instead.}%
\else
\language=\csname l@#1\endcsname
\fi
#2}}

\bibitem{he2025asap}
T.~He, J.~Gao, \emph{et~al.}, ``{ASAP}: Aligning simulation and real-world
  physics for learning agile humanoid whole-body skills,'' in \emph{Proc. of
  Robotics: Science and Systems}, 2025.

\bibitem{haarnoja2024soccer}
T.~Haarnoja, B.~Moran, \emph{et~al.}, ``Learning agile soccer skills for a
  bipedal robot with deep reinforcement learning,'' \emph{Science Robotics},
  2024.

\bibitem{zhang2026latent}
Z.~Zhang, H.~Lu, \emph{et~al.}, ``Learning athletic humanoid tennis skills from
  imperfect human motion data,'' \emph{arXiv preprint arXiv:2603.12686}, 2026.

\bibitem{huang2026adapt}
T.~Huang, R.~Liu, \emph{et~al.}, ``Towards professional tennis styles for
  humanoid robots with adaptive motion planning and tracking,'' \emph{arXiv
  preprint arXiv:2608.20087}, 2026.

\bibitem{zaidi2023esther}
Z.~Zaidi, D.~Martin, \emph{et~al.}, ``Athletic mobile manipulator system for
  robotic wheelchair tennis,'' \emph{IEEE RA-L}, 2023.

\bibitem{chen2024spatialvlm}
B.~Chen, Z.~Xu, \emph{et~al.}, ``{SpatialVLM}: Endowing vision-language models
  with spatial reasoning capabilities,'' in \emph{Proc. of the IEEE/CVF
  Conference on Computer Vision and Pattern Recognition (CVPR)}, 2024.

\bibitem{zhou2025vlm4d}
S.~Zhou, A.~Vilesov, \emph{et~al.}, ``{VLM4D}: Towards spatiotemporal awareness
  in vision language models,'' in \emph{ICCV}, 2025.

\bibitem{cheng2024spatialrgpt}
A.-C. Cheng, H.~Yin, \emph{et~al.}, ``{SpatialRGPT}: Grounded spatial reasoning
  in vision-language models,'' in \emph{Advances in Neural Information
  Processing Systems}, vol.~37, 2024.

\bibitem{grauman2022ego4d}
K.~Grauman, A.~Westbury, \emph{et~al.}, ``{Ego4D}: Around the world in 3,000
  hours of egocentric video,'' in \emph{Proc. of the IEEE/CVF Conf. on Computer
  Vision and Pattern Recognition}, 2022.

\bibitem{nair2023r3m}
S.~Nair, A.~Rajeswaran, \emph{et~al.}, ``{R3M}: A universal visual
  representation for robot manipulation,'' in \emph{Proc. of the 6th Conference
  on Robot Learning}, vol. 205, 2023.

\bibitem{driess2023palme}
D.~Driess, F.~Xia, \emph{et~al.}, ``{PaLM-E}: An embodied multimodal language
  model,'' in \emph{Proc. of the 40th Int. Conf. on Machine Learning}, 2023.

\bibitem{zhou2025dinowm}
G.~Zhou, H.~Pan, \emph{et~al.}, ``{DINO-WM}: World models on pre-trained visual
  features enable zero-shot planning,'' in \emph{Proc. of the 42nd Int.
  Conference on Machine Learning}, vol. 267, 2025.

\bibitem{assran2025vjepa2}
M.~Assran, A.~Bardes, \emph{et~al.}, ``{V-JEPA 2}: Self-supervised video models
  enable understanding, prediction and planning,'' \emph{arXiv preprint
  arXiv:2506.09985}, 2025.

\bibitem{majumdar2023vc1}
A.~Majumdar, K.~Yadav, \emph{et~al.}, ``Where are we in the search for an
  artificial visual cortex for embodied intelligence?'' in \emph{NeurIPS},
  2023.

\bibitem{alberico2025egopingpong}
I.~Alberico, M.~Cannici, \emph{et~al.}, ``Egocentric event-based vision for
  ping pong ball trajectory prediction,'' in \emph{Proc. of the IEEE/CVF Conf.
  on Computer Vision and Pattern Recognition Workshops (CVPRW)}, 2025.

\bibitem{ren2026smash}
J.~Ren, Y.~Li, \emph{et~al.}, ``{SMASH}: Mastering scalable whole-body skills
  for humanoid ping-pong with egocentric vision,'' \emph{arXiv preprint
  arXiv:2604.01158}, 2026.

\bibitem{xiao2024ball}
Q.~Xiao, Z.~Zaidi, and M.~Gombolay, ``Multi-camera asynchronous ball
  localization and trajectory prediction with factor graphs and human poses,''
  in \emph{IEEE Int. Conf. on Robotics and Automation (ICRA)}, 2024.

\bibitem{huang2019tracknet}
Y.-C. Huang, I.-N. Liao, \emph{et~al.}, ``{TrackNet}: A deep learning network
  for tracking high-speed and tiny objects in sports applications,'' in
  \emph{IEEE Int. Conf. on Advanced Video and Signal Based Surveillance}, 2019.

\bibitem{dong2026racketvision}
L.~Dong, Y.~Yang, \emph{et~al.}, ``{RacketVision}: A multiple racket sports
  benchmark for unified ball and racket analysis,'' \emph{AAAI}, 2026.

\bibitem{zhang2021vid2player}
H.~Zhang, C.~Sciutto, \emph{et~al.}, ``{Vid2Player}: Controllable video sprites
  that behave and appear like professional tennis players,'' \emph{ACM
  Transactions on Graphics}, vol.~40, 2021.

\bibitem{zhang2023vid2player3d}
H.~Zhang, Y.~Yuan, \emph{et~al.}, ``Learning physically simulated tennis skills
  from broadcast videos,'' \emph{ACM Transactions on Graphics}, vol.~42, 2023.

\bibitem{yang2026courtsi}
Y.~Yang, Y.~Shao, \emph{et~al.}, ``Stepping {VLMs} onto the court: Benchmarking
  spatial intelligence in sports,'' \emph{arXiv preprint arXiv:2603.09896},
  2026.

\bibitem{lee2024arsports}
J.~Lee, Y.~Li, \emph{et~al.}, ``Towards {AI}-powered {AR} for enhancing sports
  playability for people with low vision: An exploration of {ARSports},'' in
  \emph{Int. Symposium on Mixed and Augmented Reality Adjunct}, 2024.

\bibitem{xue2023ae2}
Z.~Xue and K.~Grauman, ``Learning fine-grained view-invariant representations
  from unpaired ego-exo videos via temporal alignment,'' in \emph{NeurIPS},
  2023.

\bibitem{khirodkar2023egohumans}
R.~Khirodkar, A.~Bansal, \emph{et~al.}, ``{Ego-Humans}: An ego-centric {3D}
  multi-human benchmark,'' in \emph{Int. Conf. on Computer Vision}, 2023.

\bibitem{shinkai2022gaze}
R.~Shinkai, S.~Ando, \emph{et~al.}, ``Visual strategies for eye and head
  movements during table tennis rallies,'' \emph{Frontiers in Sports and Active
  Living}, vol.~4, 2022.

\bibitem{tevet2023human}
G.~Tevet, S.~Raab, \emph{et~al.}, ``Human motion diffusion model,'' in
  \emph{Int. Conf. on Learning Representations}, 2023.

\bibitem{liquidai2025lfm2}
{Liquid AI}, ``Lfm2 technical report,'' \emph{arXiv preprint arXiv:2511.23404},
  2025.

\bibitem{bradski2000opencv}
G.~Bradski, ``The {OpenCV} library,'' \emph{Dr. Dobb's Journal of Software
  Tools}, vol.~25, 2000.

\bibitem{tarashima2023wasb}
S.~Tarashima, M.~A. Haq, \emph{et~al.}, ``Widely applicable strong baseline for
  sports ball detection and tracking,'' in \emph{British Machine Vision Conf.
  (BMVC)}, 2023.

\bibitem{yolo11}
G.~Jocher and J.~Qiu, ``Ultralytics {YOLO11},''
  \url{https://github.com/ultralytics/ultralytics}, 2024, version 11.0.0.

\bibitem{simeoni2026dinov}
O.~Sim{\'e}oni, H.~V. Vo, \emph{et~al.}, ``{DINO}v3,'' \emph{Transactions on
  Machine Learning Research}, 2026, featured Certification.

\bibitem{tschannen2025siglip}
M.~Tschannen, A.~Gritsenko, \emph{et~al.}, ``Siglip 2: Multilingual
  vision-language encoders with improved semantic understanding, localization,
  and dense features,'' \emph{arXiv preprint arXiv:2502.14786}, 2025.

\bibitem{lin2026depth}
H.~Lin, S.~Chen, \emph{et~al.}, ``Depth anything 3: Recovering the visual space
  from any views,'' in \emph{Int. Conf. Learning Representations}, 2026.

\bibitem{carion2026sam}
N.~Carion, L.~Gustafson, \emph{et~al.}, ``Sam 3: Segment anything with
  concepts,'' in \emph{Int. Conf. on learning representations}, vol. 2026,
  2026.

\bibitem{bolya2025PerceptionEncoder}
D.~Bolya, P.-Y. Huang, \emph{et~al.}, ``Perception encoder: The best visual
  embeddings are not at the output of the network,'' in \emph{Advances in
  Neural Information Processing Systems}, vol.~38, 2025.

\bibitem{dosovitskiy2020vit}
A.~Dosovitskiy, L.~Beyer, \emph{et~al.}, ``An image is worth 16x16 words:
  Transformers for image recognition at scale,'' \emph{ICLR}, 2021.

\bibitem{qwen3vl2025}
{Qwen Team}, ``{Qwen3-VL} technical report,'' \emph{arXiv preprint
  arXiv:2511.21631}, 2025.

\bibitem{cho2025PerceptionLM}
J.~H. Cho, A.~Madotto, \emph{et~al.}, ``{PerceptionLM}: Open-access data and
  models for detailed visual understanding,'' in \emph{Advances in Neural
  Information Processing Systems}, vol.~38, 2025.

\bibitem{hu2021lora}
E.~J. Hu, Y.~Shen, \emph{et~al.}, ``Lora: Low-rank adaptation of large language
  models,'' \emph{arXiv preprint arXiv:2106.09685}, 2021.

\bibitem{reuss2025flower}
M.~Reuss, H.~Zhou, \emph{et~al.}, ``{FLOWER}: Democratizing generalist robot
  policies with efficient vision-language-flow models,'' in \emph{9th Annual
  Conference on Robot Learning}, 2025.

\bibitem{perez2018film}
E.~Perez, F.~Strub, \emph{et~al.}, ``Film: Visual reasoning with a general
  conditioning layer,'' in \emph{AAAI}, vol.~32, no.~1, 2018.

\bibitem{shafir2024prior}
Y.~Shafir, G.~Tevet, \emph{et~al.}, ``Human motion diffusion as a generative
  prior,'' in \emph{Int. Conf. on Learning Representations}, 2024.

\bibitem{zhao2025dartcontrol}
K.~Zhao, G.~Li, and S.~Tang, ``{DartControl}: A diffusion-based autoregressive
  motion model for real-time text-driven motion control,'' in \emph{Int. Conf.
  on Learning Representations}, 2025.

\end{thebibliography}
